\documentclass{article}

\usepackage{microtype}
\usepackage{graphicx}
\usepackage{subfigure}
\usepackage{booktabs} 
\usepackage{enumitem}
\setlist{nosep}
\usepackage{url}

\usepackage{array}
\usepackage{multirow}
\usepackage{xcolor}
\usepackage[frozencache,cachedir=.]{minted}

\newcolumntype{P}[1]{>{\centering\arraybackslash}p{#1}}

\usepackage{hyperref}

\usepackage[accepted]{mlsys2023}

\mlsystitlerunning{MegaBlocks: Efficient Sparse Training with Mixture-of-Experts}

\begin{document}

\twocolumn[
\mlsystitle{MegaBlocks: Efficient Sparse Training with Mixture-of-Experts}



\mlsyssetsymbol{equal}{*}

\begin{mlsysauthorlist}
\mlsysauthor{Trevor Gale}{stanford}
\mlsysauthor{Deepak Narayanan}{msr}
\mlsysauthor{Cliff Young}{google}
\mlsysauthor{Matei Zaharia}{stanford}
\end{mlsysauthorlist}

\mlsysaffiliation{stanford}{Stanford University, Stanford, California, USA}
\mlsysaffiliation{msr}{Microsoft Research, Redmond, Washington, USA}
\mlsysaffiliation{google}{Google Research, Mountain View, California, USA}

\mlsyscorrespondingauthor{Trevor Gale}{tgale@cs.stanford.edu}

\mlsyskeywords{Deep Learning, Sparsity, Mixture-of-Experts, Transformers}

\vskip 0.3in

\begin{abstract}
We present MegaBlocks, a system for efficient Mixture-of-Experts (MoE) training on GPUs. Our system is motivated by the limitations of current frameworks, which restrict the dynamic routing in MoE layers to satisfy the constraints of existing software and hardware. These formulations force a tradeoff between model quality and hardware efficiency, as users must choose between dropping tokens from the computation or wasting computation and memory on padding. To address these limitations, we reformulate MoE computation in terms of block-sparse operations and develop new block-sparse GPU kernels that efficiently handle the dynamism present in MoEs. Our approach never drops tokens and maps efficiently to modern hardware, enabling end-to-end training speedups of up to \textbf{40\%} over MoEs trained with the state-of-the-art Tutel library and \textbf{2.4$\times$} over DNNs trained with the highly-optimized Megatron-LM framework.
\end{abstract}
]



\printAffiliationsAndNotice{}  

\section{Introduction}
\label{sec::introduction}

\begin{figure*}[ht!]
  \includegraphics[width=\textwidth]{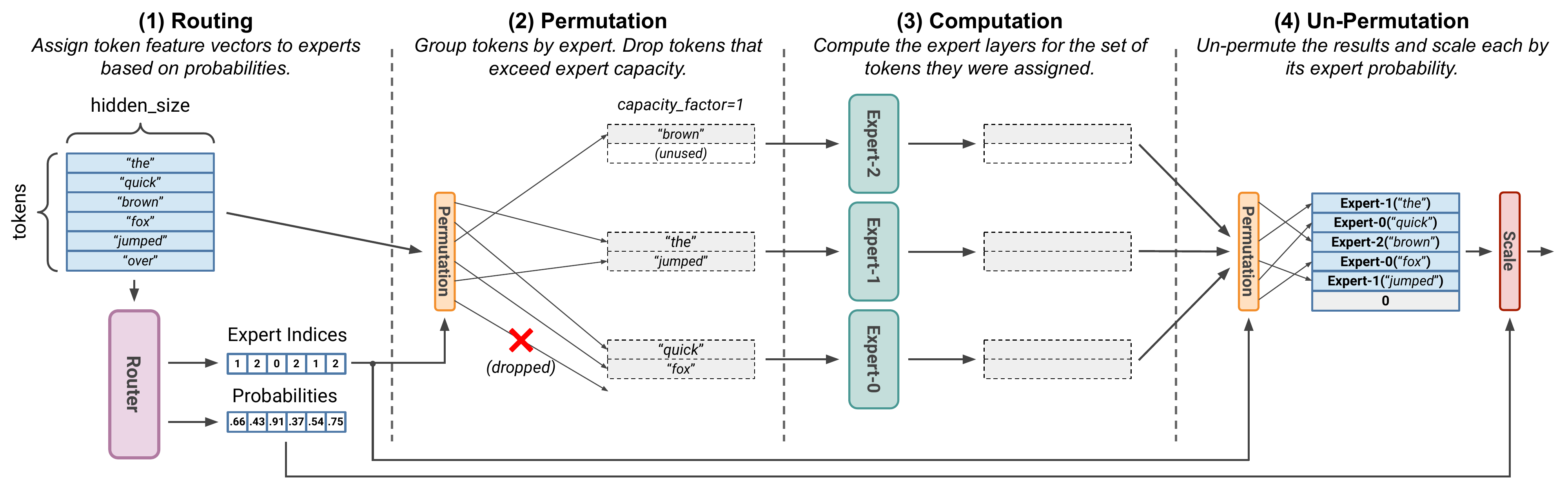}
  \vspace{-8mm}
  \caption{\textbf{A Mixture-of-Experts Layer.} Shown for \textit{num\_experts=3}, \textit{top\_k=1} and \textit{capacity\_factor=1} with the prevalent, token dropping formulation. \textbf{First (1)}, tokens are mapped to experts by the router. Along with expert assignments, the router produces probabilities that reflect the confidence of the assignments. \textbf{Second (2)}, the feature vectors are permuted to group tokens by expert assignment. If the number of tokens assigned to an expert exceeds its capacity, extra tokens are dropped. \textbf{Third (3)}, the expert layers are computed for the set of tokens they were assigned as well as any padding needed for unused capacity. \textbf{Lastly (4)}, the results of the expert computation are un-permuted and weighted by the router probabilities. The outputs for dropped tokens are shown here set to zero.}
\label{fig::moe-explained}
\end{figure*}

Exploiting sparsity in the weights, activations and input data of deep neural networks (DNNs) is an effective technique for reducing the amount of computation that is needed to achieve a given model quality \cite{lwac, tsos}. 
The past decade has seen significant progress in algorithms and high-performance software to make sparsity practically useful  \cite{blocksparse-gpu-kernels, blocksparse-rnn, wavernn, fast-sparse-convnets, sgk}.
One area that remains a challenge for sparsity is model training on accelerators. DNNs are most commonly trained on hardware accelerators like GPUs \cite{ampere-whitepaper} and TPUs \cite{google-tpu}, which exploit the regularity of dense computation to deliver high performance. Consequently, fine-grained sparse computation is less efficient on these processors. 
To enable efficient computation on accelerators, structure can be enforced on the sparse matrices \cite{blocksparse-rnn, blocksparse-gpu-kernels, balanced-sparsity}. 

%

An emerging class of models with underlying structured sparsity is Mixture-of-Experts (MoEs) \cite{sgmoe}. Each layer in an MoE is a collection of \textit{experts}, which are themselves small DNNs. As data is passed through the MoE layers, each token is dynamically routed to a subset of the experts for computation. By exploiting this sparse computation, MoEs have reduced training times by as much as 4$\times$ for applications in natural language processing and computer vision \cite{facebook-moe, vision-moe}. These gains have translated to new levels of scale for model training, pushing model sizes past 1 trillion parameters \cite{facebook-moe, google-glam, switch-transformer}.

The challenge in computing MoEs efficiently is handling the dynamic routing and load-imbalanced computation that are fundamental to these architectures. However, existing hardware and software for deep learning make it difficult to meet this challenge. For example, TPUs and their XLA compiler require all tensor shapes to be known statically and often struggle with fine-grained operations like scatters and gathers \cite{switch-transformer}. These constraints make it difficult to implement MoEs directly on TPUs. While GPUs are more flexible, the sparse computation in MoEs does not map cleanly to the software primitives supported in major frameworks and libraries. 

State-of-the-art frameworks for MoE training sidestep these challenges by placing rigid constraints on MoE routing. In order to remove the dynamism from the computation, the set of tokens mapped to each expert are trimmed or padded to a user-specified size \cite{gshard, switch-transformer, tutel}. This procrustean formulation introduces a tradeoff between model quality and hardware efficiency, as users must decide whether to drop tokens or waste computation and memory on padding. This decision is often made through hyperparameter tuning, which increases the complexity of using MoEs.

To address these challenges, we develop an approach for MoE routing and computation \textit{based on sparse primitives}. Our approach never drops tokens and maps efficiently to modern GPUs, enabling end-to-end training speedups of up to \textbf{40\%} and \textbf{2.4$\times$} over state-of-the-art frameworks for MoE and DNN training, respectively. We make the following specific contributions:
\begin{itemize}
    \item We show how the computation in an MoE layer can be expressed as block-sparse operations to accommodate imbalanced assignment of tokens to experts. We use this formulation to train \textit{dropless-MoEs} (dMoEs).
    \item We develop high-performance GPU kernels for block-sparse matrix products that efficiently handle dynamic MoE computation. Our kernels use two techniques, \textit{blocked-CSR-COO} encoding and \textit{transpose indices}, to enable efficient matrix products with sparse inputs and outputs in transposed or non-transposed order.
\end{itemize}

We have implemented these techniques in a system called MegaBlocks, which builds on the state-of-the-art Megatron-LM library for training Transformer models \cite{megatron-lm}. We evaluate our system through both microbenchmarks and end-to-end training of Transformer language models. 

\section{Background: MoE Layers}
\label{sec:::moe-layers}

MoE layers are made up of many \textit{experts}, which are themselves small neural networks. Each token\footnote{For natural language, data is commonly called \textit{tokens}. For vision, the data is typically \textit{pixels} or \textit{patches} \cite{vit}. For simplicity, we use the term token throughput this paper.} is dynamically routed to a subset of the experts for computation based on scores computed by a \textit{router}. The experts are commonly defined to be small multi-layer perceptrons (MLPs). It is typical for tokens to be sent to a small number of experts, often between 1 and 4 \cite{switch-transformer}. 

MoE layers are often interleaved with other DNN layers and are most commonly used to replace the feed-forward network (FFN) layers in Transformers \cite{sgmoe, switch-transformer}. This hybrid architecture has demonstrated strong results on both natural language and vision tasks \cite{google-glam, vision-moe}. It is conjectured that these improvements are a result of experts specializing to different parts of the data distribution \cite{sgmoe}. 


We illustrate an MoE layer in Figure \ref{fig::moe-explained} and describe it in detail in the remainder of this section.

\subsection{Routing}

The first stage of an MoE layer is the router, which is responsible for determining the assignment of tokens to experts. In addition to expert assignments, MoE routers also produce probabilities for each assignment that reflect the confidence of the mapping. These weights are encoded as a matrix of scores for each token-expert pair, which are used to linearly combine the \textit{top\_k} expert outputs for each token (see \S \ref{sec::unpermute}).

The most common style of MoE routing is the learned router proposed by \citet{sgmoe}. In this router, the tokens are projected from \textit{hidden\_size} elements to \textit{num\_experts} scores by multiplying with a weight matrix that is learned jointly with the other model parameters. The scores are normalized with a softmax and the routing decisions are made by greedily selecting the \textit{top\_k} scoring experts for each token.

\begin{figure}[t!]
  \includegraphics[width=\columnwidth]{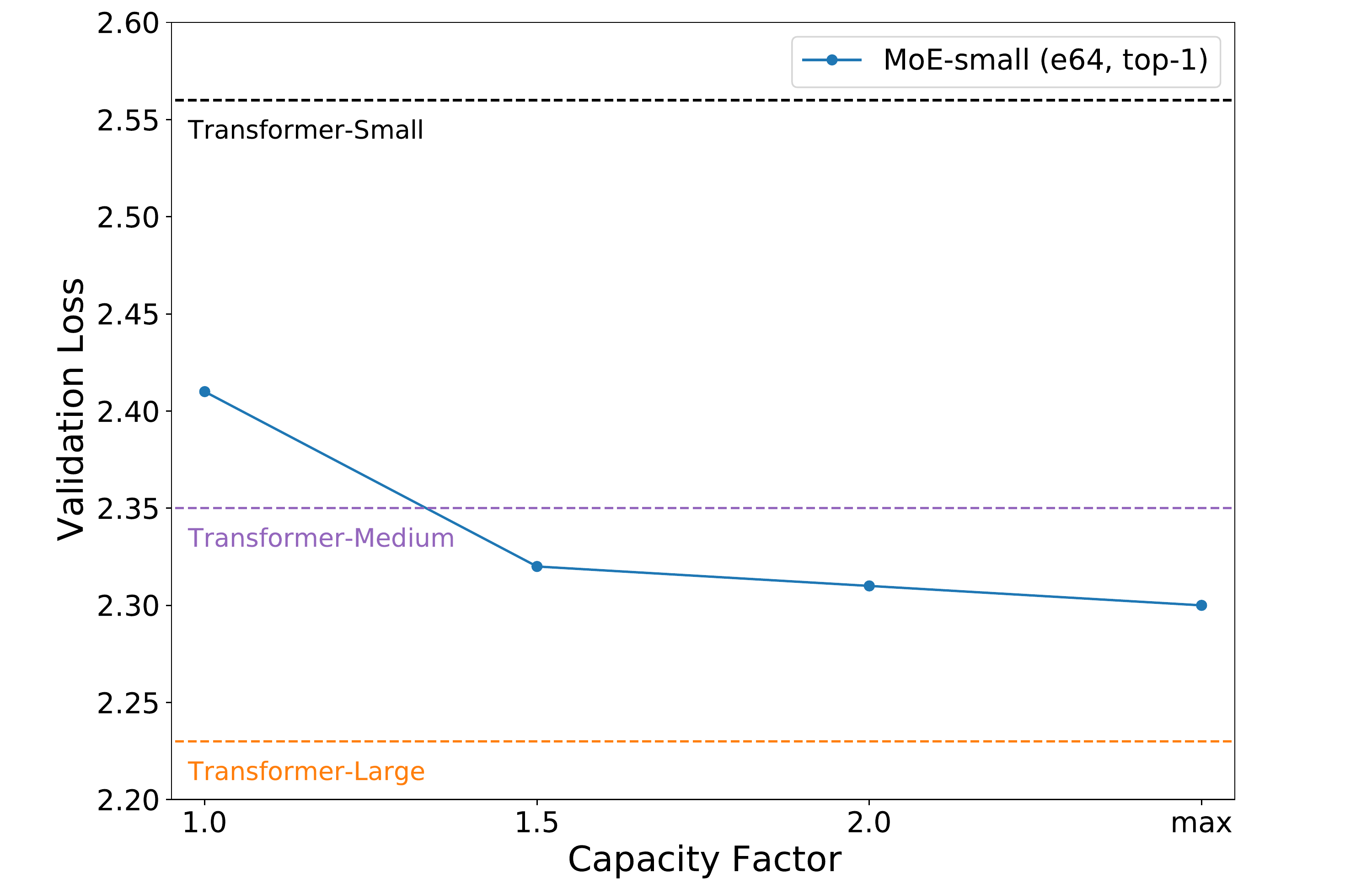}
  \vspace{-8mm}
  \caption{\textbf{MoEs Trained on The Pile with Different Capacity Factors.} The loss reached by the MoE models decreases significantly as expert capacity is increased, but at the cost of additional computation. The lowest loss is achieved by the “max” capacity factor model, which avoids dropping tokens through the dynamic capacity factor mechanism proposed by \citet{tutel}.}
\label{fig::token-dropping}
\end{figure}

\subsection{Permutation}
\label{sec::permutation}

\begin{figure*}[ht!]
  \includegraphics[width=\textwidth]{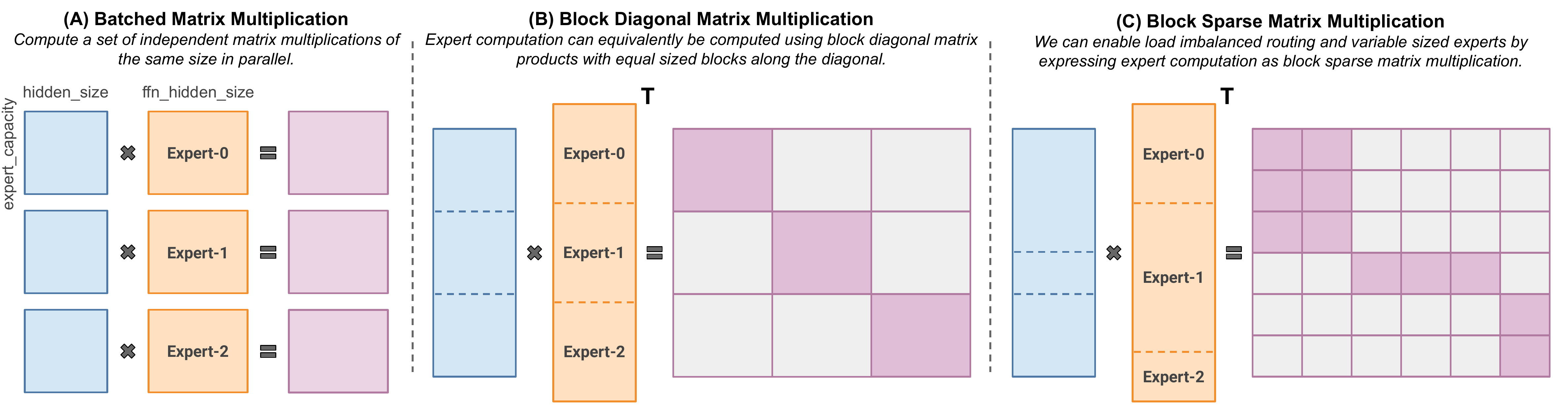}
  \vspace{-5mm}
  \caption{\textbf{Expert Computation in an MoE Layer.} Shown with \textit{num\_expert=3}. \textbf{(A)} State-of-the-art MoE implementations use batched matrix multiplication to compute all experts within a layer in parallel. This introduces the constraints that all experts are assigned the same number of tokens and that all experts have the same shape. \textbf{(B)} Expert computation can be analogously posed in terms of block diagonal matrix multiplication with identically sized blocks. \textbf{(C)} In order to relax these constraints, we can construct a block diagonal matrix with variable sized blocks made up of many smaller blocks. We can compute this matrix efficiently using block-sparse matrix multiplication.}
\label{fig::block-sparse-moe-explained}
\end{figure*}

State-of-the-art MoE implementations aim to compute all expert layers in parallel in order to make effective use of the parallelism available on GPUs and TPUs \cite{gshard, switch-transformer, tutel}. The standard primitive used by implementations is batched matrix multiplication, which computes a set of matrix products of the same shape (see Figure \ref{fig::block-sparse-moe-explained}A). However, mapping MoE computation to this primitive is non-trivial. In order to respect the shape constraints of batched matrix multiplication, the experts must be constrained to have weight matrices of the same shape and the number of tokens assigned to each expert must be equal. The latter constraint is particularly problematic because the learned routing algorithm described above provides no guarantees of a load balanced assignment of tokens to experts.

In order to satisfy this constraint, prior work has defined a fixed expert capacity, which is the number of tokens that each expert can be assigned (\citet{gshard, switch-transformer}). If the number of tokens assigned to an expert exceeds its capacity, the extra tokens are dropped. That is to say, they are not passed to any expert for computation and the model relies on a residual connection to reintroduce the dropped tokens' representation after the MoE layer. If an expert layer is not assigned enough tokens to fill its capacity, its set of tokens is padded to fill the remaining space. Expert capacity is typically specified in terms of a \textit{capacity\_factor} hyperparameter, which is a multiplier on the expected number of tokens that would be assigned to each expert under a perfect uniform distribution:

\[
expert\_capacity = \frac{num\_tokens}{num\_experts} \times capacity\_factor
\]

The \textit{capacity\_factor} can be thought of as a parameter that reduces the chance of dropping a token. This hyperparameter represents a tradeoff between additional computation and model quality. As such, it is desirable to minimize the amount of load imbalance in the assignment of tokens to experts. The typical mechanism for doing so is auxiliary \textit{load balancing losses}, which incentivize the router to produce a balanced assignment \cite{sgmoe, gshard, switch-transformer}. These losses additionally help to ensure that all experts see a similar number of tokens during training. This is thought to be important to avoid degenerate states where some experts are assigned zero tokens and stop receiving gradient updates \cite{expert-choice-routing}.

In addition to enabling batched computation of the expert layers, these constraints allow all tensor shapes to be known statically, which is required by TPUs and XLA.

\begin{table}[t!]
\caption{\textbf{Transformer Model Configurations.} These models are based on those used by \citet{transformer} and \citet{gpt3}.
FLOPs were calculated using the expression from \citet{deepak-megatron-lm} with a single sequence. All models use \textbf{ffn\_hidden\_size=4$\times$hidden\_size}.}
\vspace{2mm}

\centering
\resizebox{\columnwidth}{!}{%
\begin{tabular}{c|cccc}
 \textbf{Transformer} & \textbf{hidden\_size} & \textbf{num\_layers} & \textbf{Weights (M)} & \textbf{GFLOPs} \\ \hline
XS & 512 & 6 & 46 & 316 \\
Small & 768 & 12 & 125 & 879 \\
Medium & 1024 & 24 & 356 & 2487 \\
Large & 1536 & 24 & 760 & 5122 \\
XL & 2048 & 24 & 1316 & 8684
\end{tabular}}
\label{tab::transformer-configs}
\end{table}

\subsection{Computation}

Once the data has been permuted, the experts can be computed in parallel. For models where the experts are MLPs, this entails computing each layer for all experts using batched matrix multiplication. For convolutional experts, the layers can be computed with grouped convolutions.

\subsection{Un-permutation}
\label{sec::unpermute}

After the experts are computed, the resulting feature vectors are un-permuted such that their ordering matches that of the input to the layer. 
The last step in MoE computation is to scale the output tokens by the scores with which they were assigned to their respective experts. When tokens are routed to more than one expert, these weighted results are summed to produce the final layer output for each token.

\section{Motivation: Token Dropping in MoEs}
\label{sec::motivation}

Despite the use of load balancing losses, prior work has shown that token routing is still highly imbalanced \cite{tutel}. To quantify the effect of token dropping on model quality, we trained MoE language models on The Pile \cite{pile} with a range of capacity factors. We train Transformer MoEs similar to those used by \citet{switch-transformer}, where each model is a Transformer with the FFN layers replaced with 64-expert MoE layers where each expert is a 2-layer MLP matching the original FFN dimensions. We used top-1 routing and based our MoE model dimensions on the Transformer-Small model described in Table \ref{tab::transformer-configs}. All models were trained using the tokenization from GPT2 \cite{gpt2} for 10B tokens with sequence length 1024, the Adam optimizer, and the learning rate and gradient clipping settings from \citet{megatron-lm}. We trained all models on a single A100 GPU with a batch size of 512 sequences. We trained MoEs with capacity factor 1, 1.5, and 2 as well as the dynamic capacity factor technique proposed by Tutel~\cite{tutel}, where the capacity factor is set dynamically to the minimum value that would avoid token dropping. As a baseline, we trained standard Transformer models across a range of sizes. All Transformer and MoE models have vocabulary size 51200, sequence length 1024 and an attention head size of 64. Our model configurations are summarized in Table \ref{tab::transformer-configs} and the results of the experiments are shown in Figure \ref{fig::token-dropping}.

For these models, we observed that the impact of token dropping is significant. While the MoE with capacity factor of 1 achieved a 0.15 reduction in validation loss, the MoE that avoided dropping tokens provided a reduction of 0.26, \textbf{1.73$\times$} larger than the gain of the former model and enough to exceed the quality of Transformer-Medium. 

While dropping tokens reduces model quality, increasing capacity factor comes at the cost of additional computation and memory. In this example, MoE-layer math operations increased by over 2$\times$ in order to avoid dropping tokens. \citet{tutel} showed that some MoEs require capacity factors as high as 11 in order to avoid dropping tokens, and other models where the necessary capacity factor to avoid dropping tokens spiked unpredictably during training.

In addition to the computational overhead of increasing the capacity factor, having to tune an additional hyperparameter can significantly increase the number of models that need to be trained for a target task. This is particularly cumbersome for large neural networks, where the cost to train a single model can run into the hundreds of thousands of dollars \cite{mosaic-llm}. Possibly as a result of this, some large studies on MoEs have declined to explore different capacity factors at all \cite{facebook-moe, dm-routing-networks}.

\section{No-Token-Left-Behind with Block Sparsity\protect\footnote{The name No-Token-Left-Behind references the technique briefly discussed by \citet{switch-transformer}, which was an unsuccessful attempt to regain the quality lost from dropping tokens.}}

This section describes how we formulate MoE layer computation in terms of block-sparse computation in order to avoid dropping tokens. The motivation for using block-sparse primitives to express MoE computation is manifold. First, as we show below, block-sparse matrices are a natural and flexible way of describing the dynamic and load imbalanced computation in MoEs. Second, block sparsity maps efficiently to hardware accelerators built around systolic array matrix multipliers like GPUs and TPUs. Because of the coarse granularity of MoE experts, we can select a block size for our implementation that is large enough to enable the computation to realize high fractions of peak device throughput. Lastly, block-sparse kernels like matrix multiplication and convolution are general-purpose primitives that are useful across a range of applications \cite{blocksparse-rnn, blocksparse-gpu-kernels, sparse-transformer, fast-sparse-convnets}. This makes investment in high-performance kernels more practical, as work can be amortized across target tasks. We could similarly invest in variable sized batched matrix multiplication kernels, but the utility of this would be limited to MoE architectures as they are designed today.

In addition to these considerations, the block-sparse formulation of MoEs exposes a new perspective on these algorithms as a form of dynamic, structured, activation sparsity. This perspective draws parallels to much of the literature on sparse training algorithms and opens up the opportunity to further improve MoEs with insights from this adjacent field.

\textbf{Preliminaries: Sparse Matrix Product Notation.} In the remainder of this paper we often refer to matrix multiplication where one of the three matrices (the two inputs and one output) is sparse and the others are dense. We borrow the notation from Triton \cite{triton} to describe these different operations. Each operation is described with a three character string where each character is either “S” for sparse or “D” for dense. The order of characters is output, followed by the left input followed by the right input. For example, the product of two dense matrices with a sparse output is ``SDD'', which is also referred to as sampled dense-dense matrix multiplication (SDDMM). This notation is useful to distinguish operations like DSD and DDS, which are different forms of sparse matrix-dense matrix multiplication (SpMM). Superscript ``T'' indicates transposition of the input arguments. For example, SDD\textsuperscript{T} indicates an SDD where the right-hand input matrix is transposed.

\subsection{Expert Computation With Block Sparsity}
\label{sec::expert-computation-with-block-sparsity}

The key insight behind our method is shown in Figure \ref{fig::block-sparse-moe-explained}. Rather than the prevailing approach of computing the experts within an MoE layer using batched matrix multiplication, we could equivalently compute the experts as an SDD where the output sparse matrix has block diagonal structure, as shown in Figure \ref{fig::block-sparse-moe-explained}B. In this formulation, allowing for a load-imbalanced assignment of tokens to experts is analogous to allowing for the blocks in the block diagonal matrix to have a variable number of rows. To achieve this, we propose to compute each block as many smaller fixed size blocks using block-sparse matrix multiplication, as shown in Figure \ref{fig::block-sparse-moe-explained}C. To construct multi-layer experts, we can iterate between SDD and DSD operations (see Figure \ref{fig::dmoe-pseudo-code}).

In this formulation, we could also relax the constraint on the number of columns in each block to build MoE layers with variable sized experts, as is shown in Figure \ref{fig::block-sparse-moe-explained}C. While this is an interesting direction for future work, we did not explore these configurations as more research is needed to identify how this capability can be used to increase efficiency.

With sufficiently large blocks, block-sparse matrix multiplication is capable of reaching high fractions of peak throughput on modern GPUs \cite{blocksparse-gpu-kernels, cusparse-blockpsparse}. The coarse-grained sparsity in MoEs lends itself to this requirement - in Transformer models using MoE FFN layers, the number of columns in the blocks shown in Figure \ref{fig::block-sparse-moe-explained}B corresponds to \textit{ffn\_hidden\_size}, which is commonly between 1024 and 8192 \cite{transformer, gpt2, gpt3}. The number of rows in these blocks corresponds to the number of tokens assigned to each expert, which is expected to be equal to the number of tokens divided by the number of experts under a uniform distribution. This can range from a few thousand to tens of thousands of tokens per expert \cite{gshard, facebook-moe, switch-transformer}. These coarse-grained blocks are many times larger than the largest tile dimensions used for dense matrix multiplication kernels, which give us the flexibility to select a block size that can match their throughput.

\begin{figure}[t!]
  \includegraphics[width=\columnwidth]{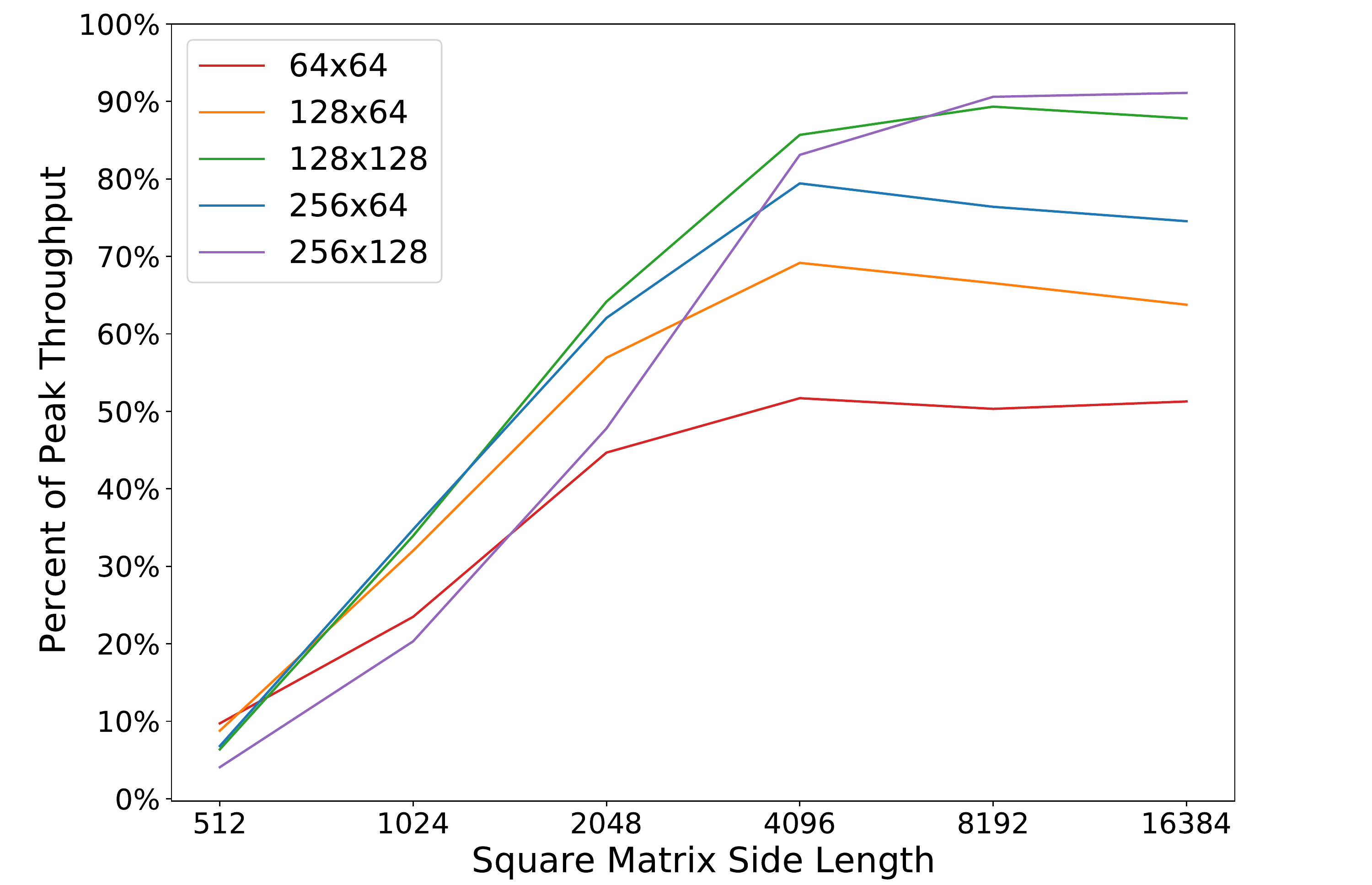}
  \caption{\textbf{Matrix Multiplication Throughput with Different Tile Dimensions.} Benchmarked on an A100 SXM4 80GB GPU with CUDA 11.5 and all tile dimensions supported by CUTLASS 2.5. We observe that 128x128 tiles perform consistently on-par or better than other configurations.}
\label{fig::cutlass-benchmarks}
\end{figure}

\section{MegaBlocks: A Framework for Efficient MoE Training}
\label{sec::implementation}

We implemented our techniques in a system called MegaBlocks, which builds on Megatron-LM \cite{megatron-lm} and PyTorch \cite{pytorch}. In addition to high-performance \textit{dropless-MoE} (dMoE) layers, our system supports distributed training of MoEs with both data and expert model parallelism \cite{switch-transformer}. 

This section discusses the design of our dMoE implementation, including our block-sparse kernels, and other considerations for building an efficient system. \S \ref{sec::existing-kernels} discusses the limitations of existing block-sparse kernels. \S \ref{sec::block-sizes} analyzes the effects of the block size on block-sparse product performance. \S \ref{sec::hybrid-format} describes our hybrid \textit{blocked-CSR-COO} sparse matrix format, which enables efficient matrix products with sparse input and output operands. \S \ref{sec::block-transpose} introduces \textit{transpose indices} as a mechanism for efficient iteration over block-sparse matrices in transposed order. Lastly, \S \ref{sec::routing-and-permutation} discusses efficient routing and permutation for dMoEs.

\textbf{Preliminaries: Matrix Multiplication on GPUs.} Matrix multiplication kernels on GPUs exploit \textit{tiling}, where the output matrix is broken up into statically sized two-dimensional blocks of values \cite{cutlass}. The computation of these tiles can be parallelized, and the individual tiles can be sized to tradeoff arithmetic intensity and parallelism. The group of threads assigned to a tile is called a \textit{threadblock}.

\subsection{Efficient Block-Sparse Kernels for MoEs}

To train MoEs with block-sparse kernels we need primitives for the forward and backward passes. Consider an MoE FFN layer where each expert is a 2-layer MLP. For this configuration, the forward pass requires an SDD operation followed by a DSD
(Figure \ref{fig::dmoe-pseudo-code}). For the backward pass, we compute SDD\textsuperscript{T} and DS\textsuperscript{T}D for the second layer data gradient and weight gradient, respectively, followed by DSD\textsuperscript{T} and DD\textsuperscript{T}S for the first layer data gradient and weight gradient, respectively. 

\subsubsection{Existing Block-Sparse Primitives}
\label{sec::existing-kernels}

We considered two existing libraries for block-sparse matrix multiplication on GPUs: NVIDIA cuSPARSE \cite{cusparse} and Triton Blocksparse \cite{triton}. cuSPARSE supports the blocked-ELL sparse matrix format for DSD. However, as of CUDA 11.8 this operation does not support transposition of the sparse matrix input. cuSPARSE also provides no SDD primitive with a blocked-ELL matrix. In addition to these limitations, the blocked-ELL format requires that all rows in the sparse matrix have the same number of nonzeros, which would defeat our goal of supporting load imbalanced matrices. Blocksparse supports SDD, DSD, and DDS as well as all combinations of transposed and non-transposed inputs. However, these primitives assume that the topology of the sparse matrices does not change between invocations\footnote{This is likely because they were written for applications like sparse attention where the sparse matrix topology is determined prior to training \cite{sparse-transformer}.}. The library API takes a bitmask describing the sparse operand and then pre-computes look-up tables and block groupings to accelerate computation. For our use case, the sparse matrix topology varies across every iteration of training and every MoE layer in the model. In order to use Blocksparse, we would have to pay the cost of these preprocessing steps repeatedly. 

Based on this analysis, we opted to write our own block-sparse primitives in order to tailor them to the dynamism of MoE expert computation. We implemented SDD, DSD, and DDS operations targeting NVIDIA GPUs. Our kernels support all combinations of transposed and non-transposed inputs. The remainder of this section details the design and implementation of our kernels.

\subsubsection{Selecting Block Size for MoEs}
\label{sec::block-sizes}

In order to efficiently use modern GPUs, we want to use sparse blocks that have sufficient arithmetic intensity to keep matrix multiplication units busy. Large blocks are also desirable to amortize the cost of storing and operating on sparse matrix metadata, since metadata like column indices only need to be kept for each block of nonzeros.

To select our target block size, we studied the performance of dense matrix multiplication kernels from NVIDIA CUTLASS \cite{cutlass} with different tile dimensions. We benchmarked mixed-precision (FP16 + FP32 accumulation) matrix multiplication on square matrices with power of two side lengths from 512 to 16384 and every set of tile dimensions supported in CUTLASS. For rectangular tiles, we show only the configurations where the first tile dimension is larger as we found these to slightly outperform the alternative ordering for these problems. We ran all benchmarks on an A100 SXM4 80GB GPU with CUDA 11.5 and CUTLASS 2.5. These benchmarks are shown in Figure \ref{fig::cutlass-benchmarks}.

Across these benchmarks, we observed that 128x128 tiles consistently perform on-par or better than other configurations. Anecdotally, we observe that this same configuration is commonly selected by NVIDIA cuBLAS \cite{cublas} for the dense Transformer models we studied. Based on this analysis, we opted to use 128x128 block sparsity. While the tile dimensions of a block-sparse matrix multiplication and the block size in the sparse matrix do not need to be equal, we found that for 128x128 blocks the highest performing tile dimensions in our workloads were also 128x128.

To implement our kernels, we extended CUTLASS \cite{cutlass} to support block-sparse matrices and reused their machinery for high-performance matrix multiplication with different data types and GPU architectures.

\begin{figure}[t!]
    \centering
  \includegraphics[width=0.92\columnwidth]{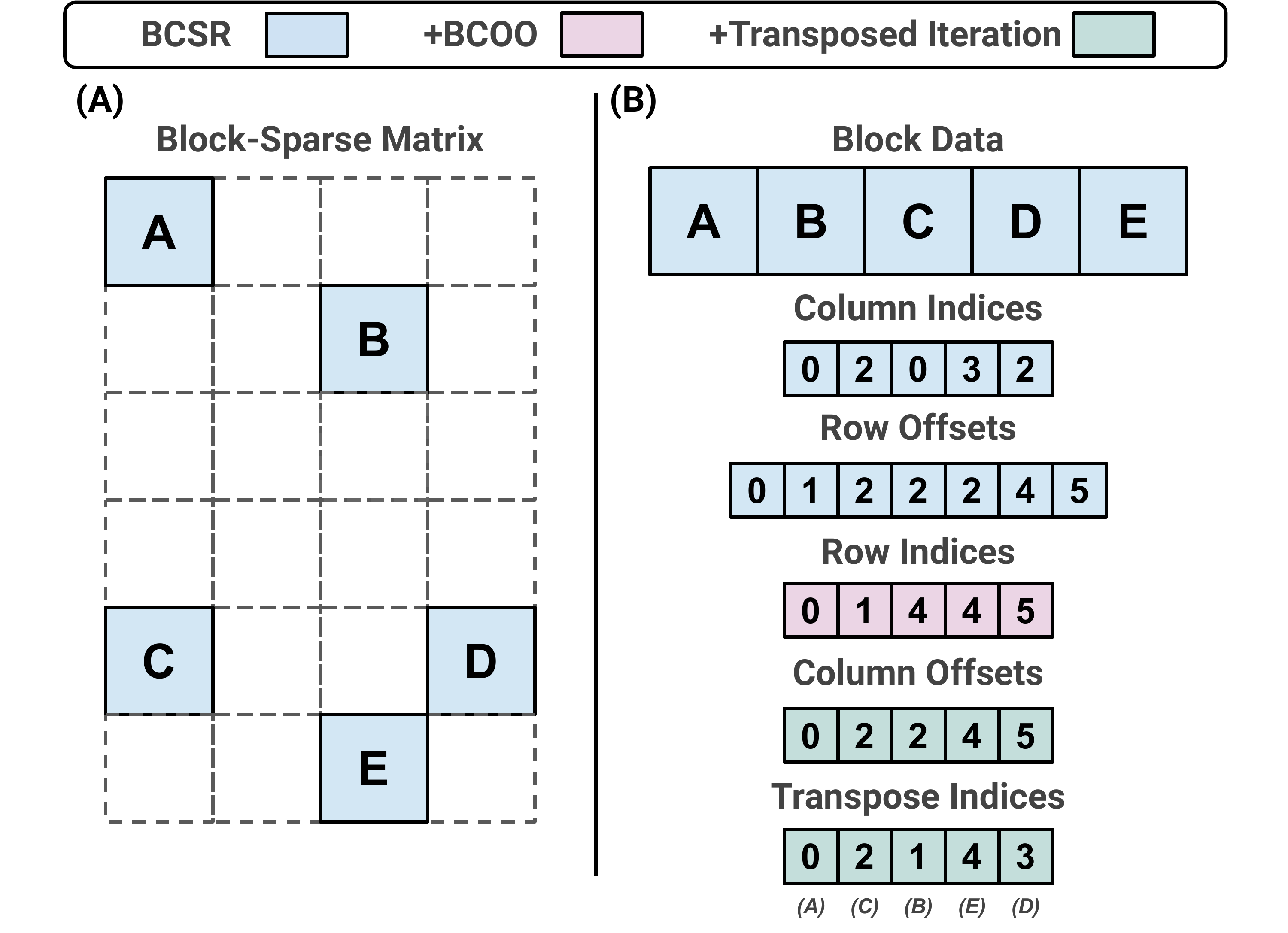}
  \caption{\textbf{Block-Sparse Matrix Format used in MegaBlocks.} Pane \textbf{(B)} shows the encoding for the sparse matrix in pane \textbf{(A)}. Indices and offsets in our encoding are block-wise. We use blocked compressed sparse row (BCSR) as our primary sparse matrix format.  We additionally store the row indices of each nonzero block (\S \ref{sec::hybrid-format}) and a secondary index of \textit{transpose indices} (\S \ref{sec::block-transpose}).}
  
\label{fig::block-sparse-format}
\end{figure}

\subsubsection{Computing Sparse Outputs With Hybrid Blocked-CSR-COO}
\label{sec::hybrid-format}

We use blocked compressed sparse row (BCSR) as our primary sparse matrix format. BCSR makes it simple to iterate across the nonzeros in a row, which is necessary for operations like DSD and DDS\textsuperscript{T}. Iterating over blocks also has minimal overhead with BCSR, as identifying a block's position in the matrix only requires a single load of its column index. We discuss our approach for efficiently iterating across the nonzeros in a column with this format in \S \ref{sec::block-transpose}.

One challenge with BCSR sparse matrices is efficiently computing SDD operations in parallel. On kernel launch, each threadblock needs to identify the row and column of its output block so that it knows which rows and columns of the input matrices are needed to compute it. Because BCSR only encodes column indices for each block, identifying the row index of a nonzero block requires a search through the row offsets. One solution to this problem is to launch the maximum number of threadblocks that could be needed to compute each row of the output if it were fully dense. On startup, each threadblock can check whether its column offset is out of range for the number of nonzeros in its row and return if there is no work to do. \citet{sgk} showed that the overhead introduced by launching extra threadblocks was negligible for moderately sparse matrices (50 - 90\% zeros). We experimented with this approach but observed that for MoEs the cost of launching these unused threadblocks was significant, particularly for models with high expert counts where the level of sparsity in the block-sparse matrices is very high.

To efficiently parallelize SDD, we additionally materialize the row indices for each nonzero block so that threadblocks can trivially look up the coordinates of sparse blocks in the output matrix. The storage required for this additional metadata is negligible since we only need to store one index per 16384 nonzero values in a 128x128 block. Even with this additional metadata, we maintain the row-wise ordering of nonzero blocks so the matrix can be operated on as either BCSR or blocked coordinate format (BCOO). We illustrate this hybrid blocked-CSR-COO encoding in Figure \ref{fig::block-sparse-format}.

\usemintedstyle{borland}
\renewcommand{\theFancyVerbLine}{\textcolor[RGB]{0,0,0}{\small \arabic{FancyVerbLine}}}
\begin{figure}[t!]
\begin{minted}[
fontfamily=courier,
fontsize=\fontsize{8pt}{8pt},
xleftmargin=14pt, 
numbersep=4pt, 
linenos, 
frame=lines]{python}
# x.shape: (num_tokens, hidden_size)
def dmoe_forward(self, x):
  # (1) Assign tokens to experts.
  #
  # indices.shape: (num_tokens)
  # weights.shape: (num_tokens)
  indices, weights = router(x)
    
  # (2) Create the sparse matrix topology.
  #
  # This describes the matrix in Figure 3C.
  topology = make_topology(indices)
    
  # (3) Permute the tokens to group by expert.
  x = padded_gather(x, indices)
    
  # (4): Compute the expert layers.
  #
  # inner_dim = ffn_hidden_size * num_experts
  # self.w1.shape: (hidden_size, inner_dim)
  # self.w1.shape: (inner_dim, hidden_size)    
  x = sdd(x, self.w1, topology)
  x = dsd(x, self.w2)
    
  # (5) Un-permute the tokens and scale.
  x = padded_scatter(x, indices)
  return x * weights
\end{minted}
\vspace{-4mm}
 \caption{\textbf{Pseudo-Code for a dMoE.} The code follows Figure \ref{fig::moe-explained} with three changes. First, we construct the sparse matrix topology from Figure \ref{fig::block-sparse-moe-explained}C from expert assignments (line 12). Second, we pad each expert batch to a multiple of the block size during permutation (line 15, \S \ref{sec::routing-and-permutation}). Lastly, we compute the experts in parallel by iterating between SDD and DSD operations (lines 22-23, \S \ref{sec::expert-computation-with-block-sparsity}).}
\label{fig::dmoe-pseudo-code}
\end{figure}

\subsubsection{Block-Sparse Transposition With Transpose Indices}
\label{sec::block-transpose}

Computing forward and backward passes for model training requires sparse matrix transposition. However, iterating over BCSR matrices in transposed order requires searching through each row to identify if the block in the target column is nonzero \cite{csb-sparse-format}. We could materialize a transposed version of the sparse matrix explicitly, but this would incur runtime and storage costs as all of the nonzero values in the matrix would need to be copied. To enable efficient iteration over BCSR matrices in transposed order, we construct the metadata for the transposed matrix but do not explicitly transpose the nonzero values. Instead, we construct an array of indices, one for each nonzero block, which are stored in transposed order and contain the offset of each nonzero block in memory. This additional metadata allows efficient iteration through the matrix in transposed order with a layer of indirection, as shown in Figure \ref{fig::block-sparse-format}. 

This idea is similar to a secondary index in a database, which allows efficient access to entries in a different order than the primary index. Similar to our hybrid Blocked-CSR-COO encoding, this technique relies on the fact that storage and computation is many times cheaper for metadata than it is for nonzero values thanks to our large block sizes.

\subsection{Efficient Routing and Permutation}
\label{sec::routing-and-permutation}

As currently implemented, our block-sparse matrix multiplication kernels require the number of tokens assigned to each expert to be a multiple of the block size. In order to respect this constraint, we pad each group of tokens with zeros to the nearest multiple of 128 and fuse this operation into custom permutation kernels. We could remove this constraint by supporting partial blocks at the fringes of the problem similar to how matrix multiplication handles matrices that are not divisible by the tile dimensions. However, the performance impact of this feature would be minimal given we expect the number of tokens assigned to each expert to be thousands or tens of thousands.

Once the expert assignments have been computed by the router, we create the metadata for the block-sparse matrix using a custom CUDA kernel. We additionally construct the transposed metadata at this time to amortize the cost over the multiple block-sparse matrix multiplications that use it across forward and backward computation.

\begin{table}[t!]
\caption{\textbf{MoE Model Configurations.} These models correspond to the Transformer configuration of the same size, but with each FFN layer replaced with a 64-expert MoE layer.}
\vspace{2mm}

\centering
\resizebox{0.95\columnwidth}{!}{%
\begin{tabular}{c|cccc}
 \textbf{MoE} & \textbf{num\_experts} & \textbf{top\_k} & \textbf{Weights (M)} & \textbf{GFLOPs} \\ \hline
XS & 64 & 1 & 839 & 316 \\
Small & 64 & 1 & 3,693 & 879 \\
Medium & 64 & 1 & 13,041 & 2487
\end{tabular}}
\label{tab::moe-configs}

\end{table}

\begin{table}[t!]
\caption{\textbf{Micro Batch Sizes Used for Model Training.} We used the largest \textit{micro\_batch\_size} that fit in memory for all experiments.}
\vspace{2mm}

\centering
\resizebox{0.9\columnwidth}{!}{%
\begin{tabular}{c|c|c}
 & \textbf{Model} & \textbf{micro\_batch\_size} \\ \hline
\multirow{5}{*}{\textbf{Megatron-LM}} & Transformer-XS & 64 \\ 
 & Transformer-Small & 32 \\
 & Transformer-Medium & 16 \\
 & Transformer-Large & 16 \\
 & Transformer-XL & 8 \\ \hline
\multirow{3}{*}{\textbf{MegaBlocks}} & dMoE-XS & 64 \\
 & dMoE-Small & 32 \\ 
 & dMoE-Medium & 8 \\ \hline
\multirow{3}{*}{\textbf{Tutel}} & dMoE-XS & 32 \\
 & dMoE-Small & 8 \\
 & dMoE-Medium & 1
\end{tabular}}
\label{tab::micro-batch-sizes}
\end{table}

\section{Experiments}

This section analyzes the performance of our system compared to state-of-the-art libraries, Microsoft Tutel \cite{tutel} and NVIDIA Megatron-LM \cite{megatron-lm}, for training Transformer MoEs and standard Transformers respectively. In order to ensure fair comparisons, we extended Megatron-LM to additionally support MoE training using Tutel’s MoE layer. All experiments were conducted on NVIDIA A100 SXM4 80GB GPUs with CUDA 11.5, CUTLASS 2.5 and used mixed-precision training \cite{mixed-precision-training} as implemented in Megatron-LM.

\begin{figure}[t!]
  \includegraphics[width=\columnwidth]{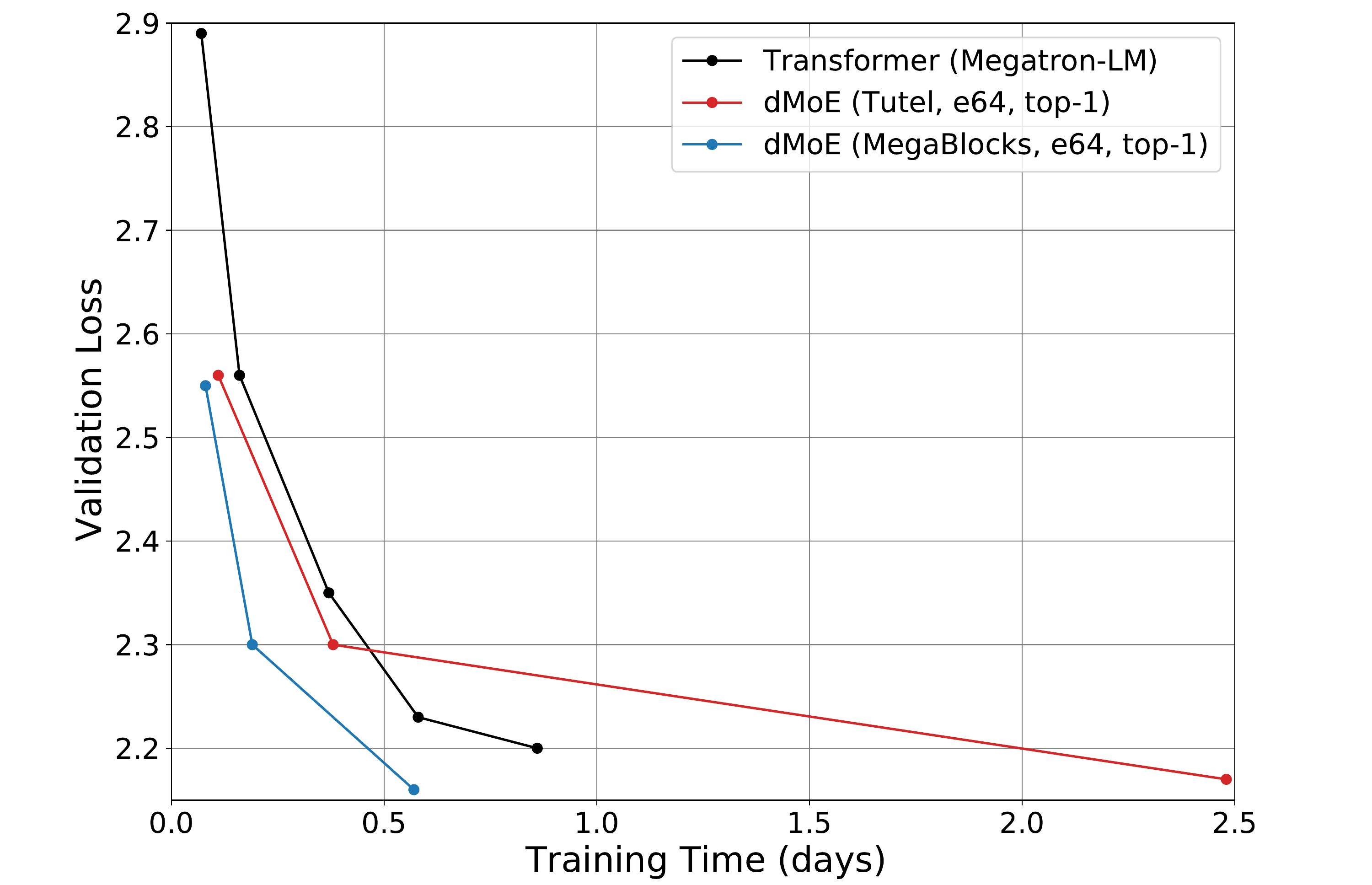}
  \vspace{-6mm}
  \caption{\textbf{MegaBlocks dMoEs, Tutel dMoEs and Megatron-LM Transformers Trained on The Pile.} MegaBlocks uses block-sparse operation to handle the dynamic and load imbalanced computation in MoEs, which enables \textbf{1.38$\times$}, \textbf{2.0$\times$} and \textbf{4.35$\times$} end-to-end training speedups for MoE-XS, MoE-Small, and MoE-Medium respectively compared to the padding-based approach used by Tutel. The advantage of our approach increases with the size of the model, as the memory requirements of padding expert batches forces Tutel to use smaller \textit{micro\_batch\_sizes} which decreases hardware efficiency. Compared to dense Transformer language models, MegaBlocks achieves \textbf{1.8$\times$} - \textbf{2.4$\times$} end-to-end training speedups for the same validation loss across these models.}
\label{fig::dropless-experiments}
\end{figure}

\subsection{MoE Training Without Dropping Tokens}

To assess the efficiency of our technique for avoiding token dropping, we compared to the dMoE method proposed by \citet{tutel} where the capacity factor is set dynamically to the minimum value that avoids token dropping.

We trained decoder-only Transformer language models on The Pile \cite{pile} with the same hyperparameters described in \S \ref{sec::motivation}. For Transformer MoEs, we trained models scaled from our XS, Small, and Medium models with each FFN layer replaced with 64-expert MoE layers using top-1 routing. We also trained standard Transformer models from 46M to 1.3B parameters, equivalent to Transformer-Base \cite{transformer} up to GPT3-XL \cite{gpt3}, as a dense baseline. We trained all models on 8 A100 SXM4 80GB GPUs using 8-way expert model parallelism for MoE layers and data parallelism for all other layers. We use gradient accumulation for all models and train with a batch size of 512 sequences and the largest \textit{micro\_batch\_size} that does not run out of memory \cite{deepak-memory-pipelining}. Our model configurations are summarized in Tables \ref{tab::transformer-configs} and \ref{tab::moe-configs}. For each model, we report the end-to-end training time and final loss achieved on a validation set in Figure \ref{fig::dropless-experiments}. 

Compared to the prevalent padding-based approach for avoiding token dropping, our technique for adaptive MoE computation with block sparsity enables end-to-end training speedups of \textbf{1.38$\times$}, \textbf{2.0$\times$} and \textbf{4.35$\times$} for MoE-XS, MoE-Small, and MoE-Medium, respectively. In addition to computational overhead, the padding-based approach implemented in Tutel significantly increases the amount of memory required to store activations in the MoE layers. This is particularly problematic because MoEs already require many times more storage for their large weight matrices compared to standard Transformers. For these models, we observed this increase in memory usage reduced the maximum \textit{micro\_batch\_size} that Tutel could use by 2$\times$, 4$\times$, and 8$\times$ compared to MegaBlocks for MoE-XS, MoE-Small, and MoE-Medium, respectively. This in turn increases training time because of reduced hardware efficiency. As a result, we observe that the advantage of MegaBlocks over Tutel grows with model size. The \textit{micro\_batch\_size} used for each model configuration are shown in Table \ref{tab::micro-batch-sizes}.

Compared to Transformer models trained with Megatron-LM, dMoEs trained with MegaBlocks reduce the training time required to reach a given validation loss by \textbf{1.8$\times$} - \textbf{2.4$\times$}. The variation in this comparison is primarily a result of the increased weight memory usage of MoE models, which forced MegaBlocks to use a 2x smaller \textit{micro\_batch\_size} for MoE-Medium than the analogous Transformer model. These results highlight the importance of reducing memory usage in MoEs as a direction for future research.

For these Transformer models, we observed that Megatron-LM sustains between 21\% and 48\% of the 2.5 petaFLOP peak throughput of this 8-GPU system with efficiency increasing with model size. The speedups achieved by MegaBlocks over this state-of-the-art framework demonstrates the efficiency of our system and the efficacy of MoEs.

\begin{figure}[t!]
  \includegraphics[width=\columnwidth]{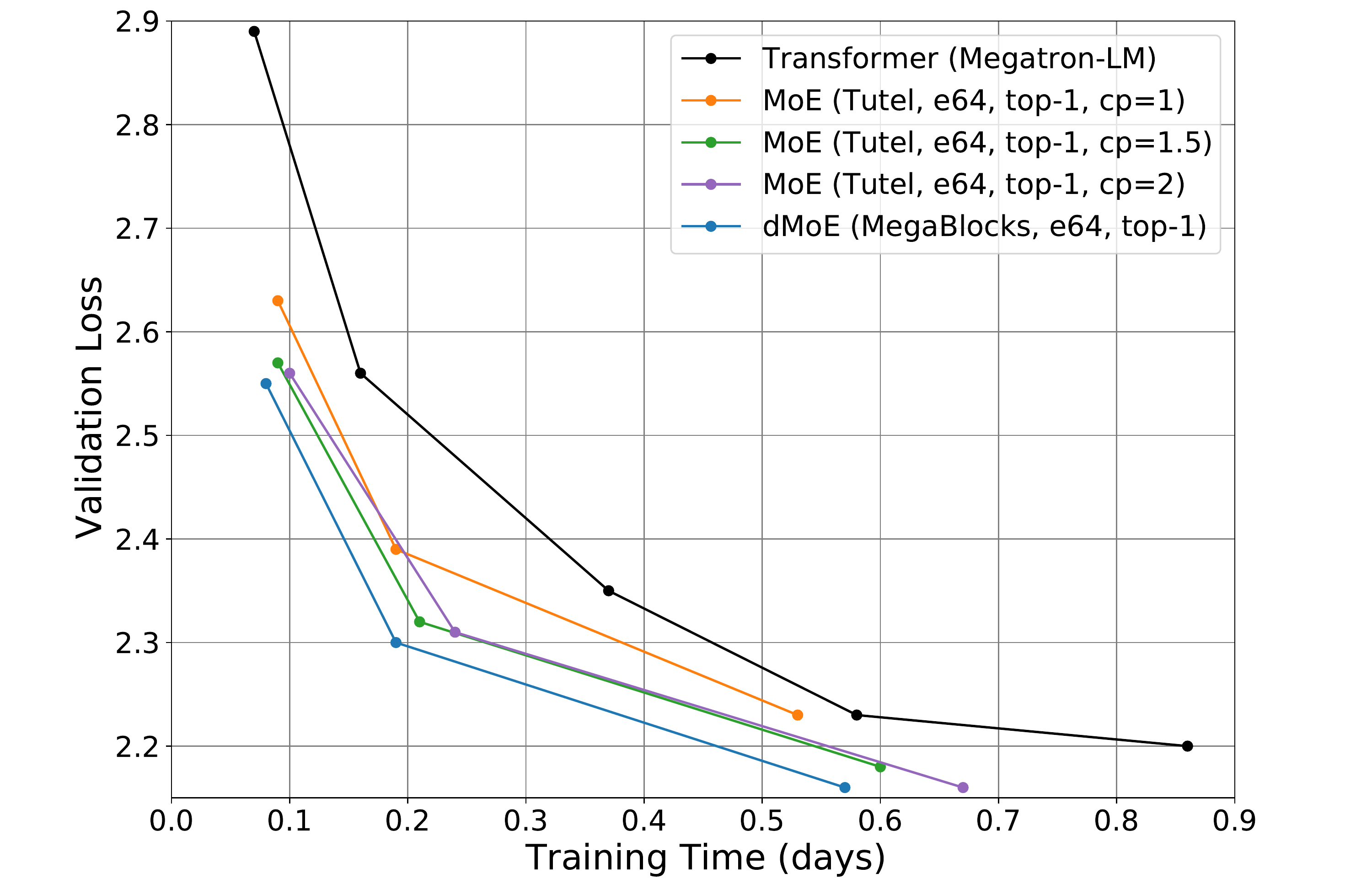}
  \vspace{-8mm}
  \caption{\textbf{MegaBlocks dMoEs, Tutel MoEs and Megatron-LM Transformers Trained on The Pile.} Even with the most efficient \textit{capacity\_factor} for each MoE, MegaBlocks reduces the training time required to reach a given validation loss by \textbf{1.38$\times$}, \textbf{1.37$\times$} and \textbf{1.18$\times$} for MoE-XS, MoE-Small and MoE-Medium respectively. In addition to these speedups, our approach reduces the cost of using MoEs by decreasing the number of hyperparameters that need to be re-tuned for each model and task.}
\label{fig::dropping-experiments}
\end{figure}

\subsection{MoE Training With Token Dropping}

We additionally compare our dMoE models to token-dropping MoEs trained with Tutel. In order to find the most efficient configurations, we trained MoE-XS, MoE-Small and MoE-Medium models with capacity factors of 1$\times$, 1.5$\times$, and 2$\times$ for a total of 9 additional models. For these configurations, all token-dropping MoE models were able to use the same \textit{micro\_batch\_size} as the analogous dMoE without running out of GPU memory. We report the end-to-end training time and validation loss for these models along with our dMoE and standard Transformer results in Figure \ref{fig::dropping-experiments}. Comparing MoEs and dMoEs for the same accuracy is non-trivial because token dropping degrades model quality. For each dMoE, we estimated the runtime of the MoE that would achieve the same validation loss by comparing to the loss-equivalent point on the MoE Pareto frontier.

\begin{figure*}[ht!]
  \includegraphics[width=\textwidth]{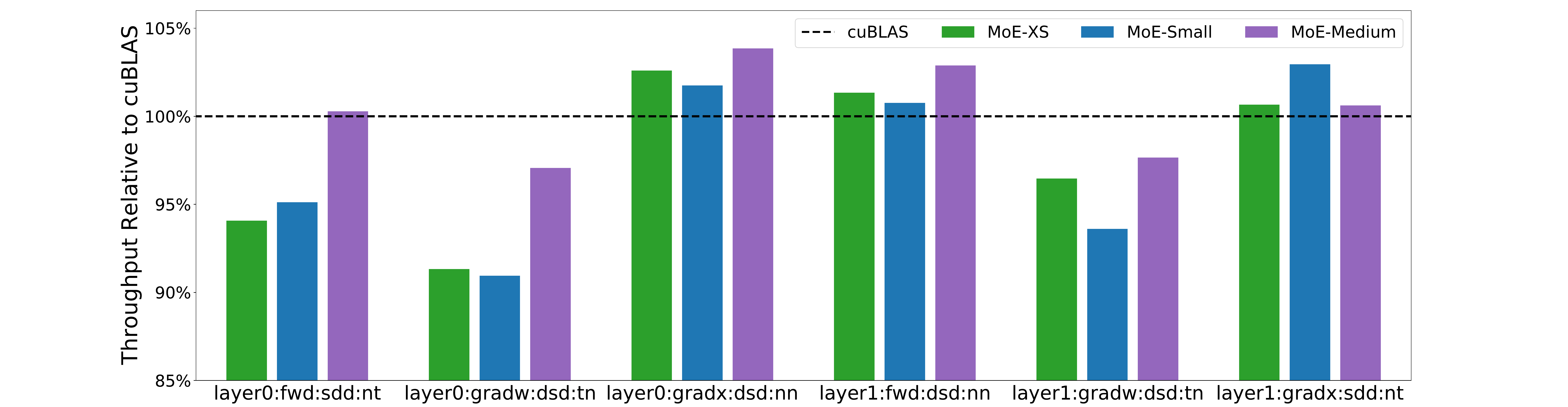}
  \vspace{-8mm}
  \caption{\textbf{Block-Sparse Matrix Multiplication Throughput Compared to cuBLAS Batched Matrix Multiplication.} Benchmarked for the problem configurations used in training MoE-XS, MoE-Small and MoE-Medium models. For these problems, our block-sparse matrix multiplication kernels realize 98.6\% of the throughput achieved by cuBLAS on average with a standard deviation of 4\% and a maximum and minimum relative throughput of 104\% and 91\% respectively.}
\label{fig::block-sparse-matmul-benchmarks}
\end{figure*}

Even with the most efficient \textit{capacity\_factor} for each MoE, dMoEs trained with MegaBlocks reduce the training time required to reach a given validation loss by \textbf{1.38$\times$}, \textbf{1.37$\times$} and \textbf{1.18$\times$} for MoE-XS, MoE-Small and MoE-Medium, respectively. In addition to significant reductions in end-to-end training time, our system reduces the cost of using MoEs by decreasing the number of hyperparameters that need to be re-tuned for each model and task. These computational savings could in turn be applied to exploring other parameters to further improve model quality.

For MoE-Medium, we observe some loss of efficiency in our implementation due to the relatively small \textit{micro\_batch\_size} that could be used while fitting in limited GPU memory. For small batch sizes, smaller tile dimensions (e.g., 64x128 or 64x64) in our block-sparse kernels could improve performance by reducing the amount of wasted computation when the problem dimensions are not divisible by 128. Another direction for increasing efficiency is to reduce the memory usage per device such that larger batch sizes can be used, either through parallelization over more devices or techniques like selective recomputation \cite{selective-recompute}.

\subsection{Block-Sparse Matrix Multiplication Performance}

To assess the quality of our block-sparse matrix multiplication kernels, we benchmarked the problem configurations used in training MoE-XS, MoE-Small and MoE-Medium models and compared to cuBLAS batched matrix multiplication. This includes the forward pass, backward weights, and backward data operations for the two layers in each FFN layer. In total, we benchmark 18 problems - 6 problems for each of the 3 models. To allow for comparison with batched matrix multiplication, we benchmarked each problem with a uniform distribution of tokens to experts and the same \textit{micro\_batch\_size} listed in Table \ref{tab::micro-batch-sizes}. These benchmarks can be viewed as an ablation assessing the overhead that would be introduced if one were to use our block-sparse kernels to implement a standard, token-dropping MoE. For each problem we averaged throughput over 100 executions. We do not include the time taken to construct the sparse matrix metadata in these benchmarks as these operations amortize over all 6 problems within an FNN layer. The results of these benchmarks are shown in Figure \ref{fig::block-sparse-matmul-benchmarks}.

On these problems, we observe that our block-sparse kernels are able to realize 98.6\% of the throughput of cuBLAS with a standard deviation of 4\%. The maximum relative throughput was 104\% and the minimum was 91\%. Overall, our kernels slightly outperformed cuBLAS on half of the problems and slightly underperformed on the other half.

While benchmarking CUTLASS, we observed that altering the order in which tiles of the output matrix are computed can change the throughput of the operation by as much as 10\% due to L2 caching effects. We believe that most of the performance discrepancy in these results can be attributed to the re-ordering of computation that occurs with block-sparse matrices, although further investigation is needed.

One case where we note additional overhead is in the DS\textsuperscript{T}D operations used to compute weight gradients. Because we use a secondary index to iterate over the sparse operand in transposed order the access patterns when iterating through this matrix exhibit little spatial locality which in turn reduces the throughput of the overall operation. While this is an interesting problem for further study, the overall impact on model performance is minimal because of the limited opportunity for improvement (\textless10\%) combined with the relatively small amount of end-to-end runtime that these two operations represent.

\section{Related Work}

\textbf{MoE Routing.} Improved routing algorithms for MoEs is an active area of research. BASE layers formulate MoE routing as a linear assignment problem trying to maximize the aggregate token-expert affinities under the constraint of a perfectly balanced assignment \cite{base-layers}. This method guarantees no tokens are dropped by re-routing tokens to different experts as needed. \citet{dm-routing-networks} found that BASE layers can incur significant runtime overhead and proposed an approximate version using the Sinkhorn algorithm. Because their approximation is no longer guaranteed to avoid token dropping, \citet{dm-routing-networks} use a capacity factor of 2 for all experiments. Other techniques have been proposed to statically decide tokens to expert mappings ahead of time based on hash functions \cite{hash-routing}. However, \citet{dm-routing-networks} observed that this approach did not perform as well as the other routing algorithms they studied. More recently, \citet{expert-choice-routing} proposed to reverse the routing problem such that each expert selects its \textit{top\_k} scoring tokens. While this guarantees a load balanced assignment of tokens to experts, this method still suffers from token dropping because the same token can be selected by multiple experts. We expect that improved routing algorithms complement our method for efficient and flexible expert computation. Exploring how these methods could be combined is an interesting direction for future research.

\textbf{High-Performance MoEs.} To scale MoE training, Tutel implements optimized distributed communication primitives for MoEs and techniques for hiding the communication costs of expert model parallelism \cite{tutel}. \citet{faster-moe} proposed FasterMoE, a system for distributed training of MoEs based on efficient communication strategies and changes to the MoE routing algorithm to avoid network congestion. Our implementation could additionally benefit from these techniques, particularly for large-scale distributed training.

\textbf{Sparse Kernels.} Sparse matrix formats that allow for efficient transposed access are well studied \cite{csb-sparse-format, csf-sparse-format, hicoo-sparse-format}. Exploring how these formats can be adapted to large block sparsity on modern GPUs is an interesting direction for future research.

\section{Conclusion}

We introduced MegaBlocks, a system for efficient MoE training on GPUs. Our system is based on a reformulation of MoEs in terms of block-sparse operations and new, block-sparse GPU kernels that efficiently handle the dynamism present in MoEs. Our approach never drops tokens and maps efficiently to modern hardware accelerators, enabling end-to-end training speedups of up to 40\% over MoEs trained with the state-of-the-art Tutel library and 2.4$\times$ over DNNs trained with the highly-optimized Megatron-LM framework.

\bibliography{main}
\bibliographystyle{mlsys2023}

%


\end{document}